\documentclass{article}

\usepackage{amsmath,amsfonts,bm}

\def\eqref#1{equation~\ref{#1}}

\def\1{\bm{1}}

\def\rmA{{\mathbf{A}}}
\def\rmB{{\mathbf{B}}}
\def\rmC{{\mathbf{C}}}

\def\rmK{{\mathbf{K}}}

\def\rmO{{\mathbf{O}}}

\def\rmQ{{\mathbf{Q}}}

\def\rmV{{\mathbf{V}}}
\def\rmW{{\mathbf{W}}}
\def\rmX{{\mathbf{X}}}

\def\vo{{\bm{o}}}

\def\vx{{\bm{x}}}
\def\vy{{\bm{y}}}

\DeclareMathAlphabet{\mathsfit}{\encodingdefault}{\sfdefault}{m}{sl}
\SetMathAlphabet{\mathsfit}{bold}{\encodingdefault}{\sfdefault}{bx}{n}

\newcommand{\R}{\mathbb{R}}

\usepackage{microtype}
\usepackage{booktabs} 

\usepackage{graphicx}
\usepackage{caption}
\usepackage{subcaption}

\usepackage{listings}
\usepackage{xcolor}

\definecolor{codegreen}{rgb}{0,0.6,0}
\definecolor{codegray}{rgb}{0.5,0.5,0.5}
\definecolor{codepurple}{rgb}{0.58,0,0.82}
\definecolor{backcolour}{rgb}{0.95,0.95,0.92}

\lstdefinestyle{mystyle}{
    backgroundcolor=\color{backcolour},   
    commentstyle=\color{codegreen},
    keywordstyle=\color{magenta},
    numberstyle=\tiny\color{codegray},
    stringstyle=\color{codepurple},
    basicstyle=\ttfamily\footnotesize,
    breakatwhitespace=false,         
    breaklines=true,                 
    captionpos=b,                    
    keepspaces=true,                 
    numbers=left,                    
    numbersep=5pt,                  
    showspaces=false,                
    showstringspaces=false,
    showtabs=false,                  
    tabsize=2
}

\usepackage{hyperref}

\usepackage[accepted]{icml2025}

\usepackage{amsmath}
\usepackage{amssymb}
\usepackage{mathtools}
\usepackage{amsthm}

\usepackage[capitalize,noabbrev]{cleveref}

\usepackage{tcolorbox}  
\usepackage{lipsum}  
\usepackage[normalem]{ulem}

\theoremstyle{plain}

\theoremstyle{definition}

\theoremstyle{remark}

\usepackage[textsize=tiny]{todonotes}

\icmltitlerunning{Thinking Slow, Fast: Scaling Inference Compute with Distilled Reasoners}

\begin{document}

\twocolumn[
\icmltitle{Thinking Slow, Fast:\\
Scaling Inference Compute with Distilled Reasoners}

\icmlsetsymbol{equal}{*}
\icmlsetsymbol{equal1}{\textdagger}

\begin{icmlauthorlist}
\icmlauthor{Daniele Paliotta}{equal,yyy,comp}
\icmlauthor{Junxiong Wang}{equal,comp,cornell}
\icmlauthor{Matteo Pagliardini}{equal,epfl}
\icmlauthor{Kevin Y. Li}{equal,cmu}
\icmlauthor{Aviv Bick}{cmu}
\icmlauthor{J. Zico Kolter}{cmu}
\icmlauthor{Albert Gu}{cmu,cartesia}
\icmlauthor{François Fleuret}{equal1,yyy,meta}
\icmlauthor{Tri Dao}{equal1,comp,sch}
\end{icmlauthorlist}

\icmlaffiliation{yyy}{Machine Learning Group, University of Geneva}
\icmlaffiliation{comp}{Together AI}
\icmlaffiliation{sch}{Princeton University}
\icmlaffiliation{cmu}{Carnegie Mellon University}
\icmlaffiliation{meta}{META}
\icmlaffiliation{epfl}{EPFL}
\icmlaffiliation{cartesia}{Cartesia.ai}
\icmlaffiliation{cornell}{Cornell University}

\icmlcorrespondingauthor{Daniele Paliotta}{daniele.paliotta@unige.ch}
\icmlkeywords{Machine Learning, ICML}

\vskip 0.3in
]

\printAffiliationsAndNotice{\icmlEqualContribution} 

 \begin{abstract}
Recent advancements have demonstrated that the performance of large language models (LLMs) can be significantly enhanced by scaling computational resources at test time. 
A common strategy involves generating multiple Chain-of-Thought (CoT) trajectories and aggregating their outputs through various selection mechanisms. 
This raises a fundamental question: can models with lower complexity leverage their superior generation throughput to outperform similarly sized Transformers for a fixed computational budget?
To address this question and overcome the lack of strong subquadratic reasoners, we distill pure and hybrid Mamba models from pretrained Transformers.
Trained on only 8 billion tokens, our distilled models show strong performance and scaling on mathematical reasoning datasets while being much faster at inference for large batches and long sequences. 
Despite the zero-shot performance hit due to distillation, both pure and hybrid Mamba models can scale their coverage and accuracy performance past their Transformer teacher models under fixed time budgets, opening a new direction for scaling inference compute.

\end{abstract}
\begin{figure*}[t!]
  \centering
  \begin{subfigure}[b]{0.575\linewidth}
  \includegraphics[width=\textwidth,clip]{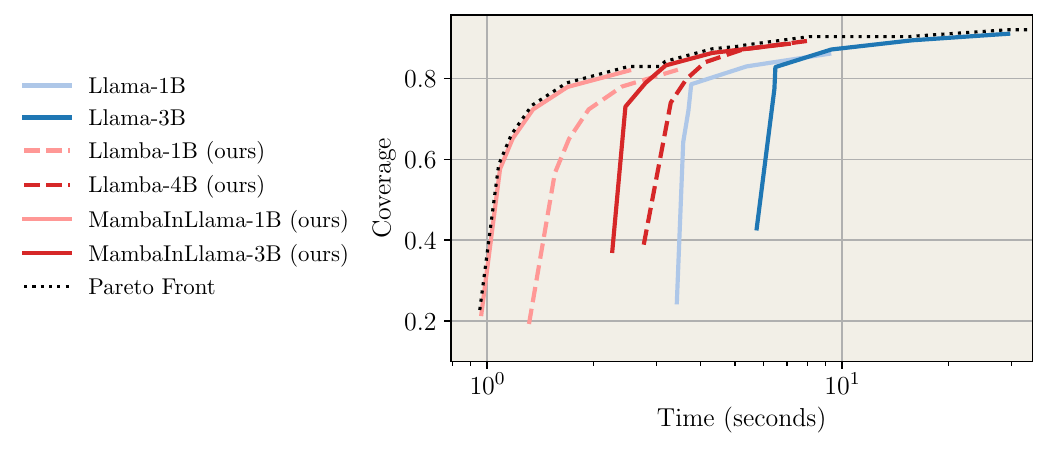}
  \caption{Scaling with time.}
  \label{fig:coverage-math-b}
  \end{subfigure}
  \begin{subfigure}[b]{0.38\linewidth}
  \includegraphics[width=\textwidth,clip]{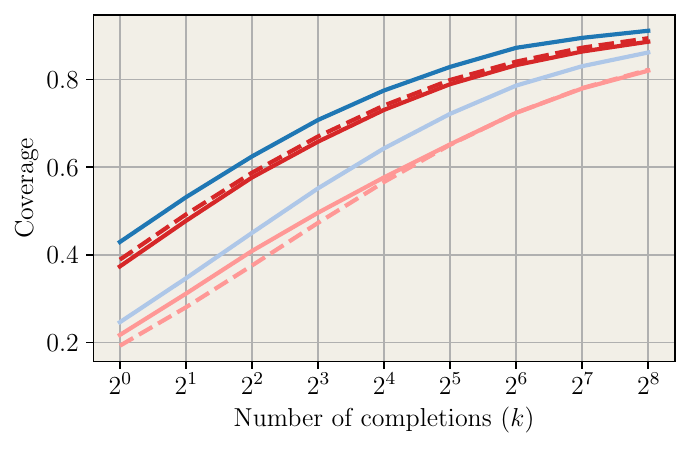}
  \caption{Scaling with number of completions.}
  \label{fig:coverage-math-a}
  \end{subfigure}
\caption{\textbf{Distilled models have better coverage on MATH for most time budgets.}
In \textbf{(b)}, we show the coverage as we increase the number of sampled answers $k$. Compared to their associated Llama baselines, our distilled models provide a lower coverage for a given $k$. In \textbf{(a)}, we now show the shortest time required to reach a given coverage. For each curve in \textbf{(b)}, we map the $k$-values on the x-axis to the time required to generate that many samples for each model. Ideally, we would want to reach the highest coverage for short time budget. For a given time budget, our distilled models can generate many more completions than their respective baselines. As such, the higher throughput of our models, shown in Figure~\ref{fig:speedup}, allows them to overcome their lower per-sample coverage. As a result, our models push the Pareto front forward for most time budgets. 
 }
\label{fig:coverage-math}
\end{figure*}

\section{Introduction}
Reasoning in large language models (LLMs) has seen a significant boost in performance recently, largely driven by scaling inference compute. 
A key technique to enhance ``reasoning'' performance is the use of intermediate reasoning steps before producing a final answer, known as Chain-of-Thought (CoT)~\citep{wei2023chainofthoughtpromptingelicitsreasoning}.
Building on this, many test-time compute techniques often involve generating multiple CoTs ~\citep{wu2024inferencescalinglawsempirical, snell2024scalingllmtesttimecompute} and selecting the best one. 
Even simple strategies, such as majority voting, can be surprisingly effective ~\citep{brown2024largelanguagemonkeysscaling, beeching2024scalingtesttimecompute}.
Furthermore, trained reward models can provide scores for the final model answers and even for the individual steps of the CoTs ~\citep{luo2024improvemathematicalreasoninglanguage}.

However, these test-time compute techniques introduce significant challenges for LLM systems. Generating long CoT sequences or large batches of completions places substantial demands on memory and compute resources. 
Transformers, in particular, struggle with such workloads due to their linear memory scaling and memory-bound nature during generation.
This raises an important question: \emph{how should we optimize model architectures to best scale test-time compute?}
In particular, can alternative architectures with faster and more efficient generation outperform current LLMs under fixed compute budgets? Addressing this problem could unlock new avenues for deploying reasoning models with different architectures, enabling them to run and scale more efficiently on hardware and environments with limited memory and compute.

Recent subquadratic architectures have training time or prefill time linear in sequence length, and constant memory requirement (instead of linear) during inference.
This enables up to 5$\times$ higher inference throughput ~\citep{gu2024mambalineartimesequencemodeling, peng2023rwkvreinventingrnnstransformer} as inference time for large batch size or long sequences is dominated by the time to load the model states (KV cache or RNN states).  
Despite their efficiency, subquadratic models have not been extensively explored in reasoning tasks, primarily due to the lack of large-scale pretrained models compared to Transformer-based counterparts. As a result, it remains unclear whether: (1) scaling inference compute for subquadratic models improves reasoning performance, and (2) subquadratic models can match or exceed Transformers models under fixed compute budgets.

In this work, we explore the reasoning capabilities of subquadratic architectures by distilling knowledge from pretrained Transformers into hybrid and pure Mamba models. 
To address the scarcity of pretrained subquadratic models with robust reasoning abilities, we develop recipes to distill specific reasoning skills into these architectures.
We then benchmark the models for multiple Chain-of-Thought (CoT) completions, providing a comprehensive analysis of performance under fixed compute and memory constraints.
Our approach advances the Pareto front established by existing models, achieving a better trade-off between efficiency and reasoning capability.

Our distilled pure and hybrid subquadratic reasoners are able to outperform their Transformer teachers on both coverage and accuracy on MATH~\citep{lightman2023letsverifystepstep} and GSM8K~\citep{cobbe2021trainingverifierssolvemath} mathematical reasoning tasks on most time budgets, reaching the same quality with 2.5$\times$ less inference time. Our results highlight the potential for distilling reasoning and math capabilities across architectures in a cost-effective manner while maintaining the inference compute scaling properties of Transformers.

\section{Related Work}
\label{sec:related}

\subsection{Scaling Inference Time Compute for Reasoning}
Scaling inference time compute has emerged as a promising strategy to improve the performance of LLMs.
Techniques such as Chain of Thought (CoT) and its variants have demonstrated significant performance improvements across various reasoning benchmarks by decomposing complex tasks into intermediate steps~\citep{wei2023chainofthoughtpromptingelicitsreasoning, yao2023treethoughtsdeliberateproblem} 

While these approaches improve reasoning through task decomposition, they also increase computational demands due to longer generation sequences. Recent work suggests that this additional compute may itself contribute to improved model abilities~\citep{pfau2024letsthinkdotdot}. Dynamic compute allocation during inference has further advanced this paradigm. For instance, \citet{goyal2024thinkspeaktraininglanguage} introduced pause tokens into the LLM vocabulary, enabling models to allocate compute more effectively and achieve better reasoning and task performance.

Another prominent approach involves generating and searching through multiple model outputs to select the best answer. Various sampling algorithms have been proposed to increase the diversity and quality of generated outputs to increase the likelihood of the correct or best answer being selected~\citep{wang2023selfconsistencyimproveschainthought, renze2024effectsamplingtemperatureproblem, zhang2023planninglargelanguagemodels}. In parallel, outcome and process reward models (ORMs and PRMs) have been introduced to help evaluate the best response and guide intermediate generation steps within the LLM model~\citep{lightman2023letsverifystepstep, zhang2024restmctsllmselftrainingprocess, luo2024improvemathematicalreasoninglanguage, uesato2022solvingmathwordproblems}.

Recent work has shown that smaller LLMs, when scaled through inference-time compute (e.g., via majority voting or PRM-guided search), can outperform larger models under fixed compute budgets~\citep{snell2024scalingllmtesttimecompute, wu2024inferencescalinglawsempirical, beeching2024scalingtesttimecompute}. However, these findings are primarily limited to Transformer-based architectures. The extent to which these scaling laws apply to subquadratic architectures, which offer faster inference but may trade off expressiveness, remains underexplored.

\subsection{Subquadratic Architecture Alternatives}
While Transformers dominate the landscape of reasoning models~\citep{grattafiori2024llama3herdmodels, qwen2025qwen25technicalreport}, alternative architectures have been proposed to mitigate their high computational cost.
These models, based on RNNs~\citep{beck2024xlstmextendedlongshortterm, peng2023rwkvreinventingrnnstransformer}, SSMs~\citep{gu2022efficientlymodelinglongsequences, gu2024mambalineartimesequencemodeling}, and linear attention mechanisms~\citep{katharopoulos2020transformersrnnsfastautoregressive, yang2024gatedlinearattentiontransformers}, offer improved inference and memory efficiency especially for long-context tasks and large-batch generation, making them attractive for large-scale language modeling. 
Notably, the Mamba family of models (Mamba-1 and Mamba-2) has introduced selective state spaces, enabling linear-time sequence modeling without sacrificing performance~\citep{gu2024mambalineartimesequencemodeling, dao2024transformersssmsgeneralizedmodels}. Hybrid architectures that combine subquadratic layers (e.g., Mamba) with a limited number of self-attention layers have also emerged, achieving superior performance compared to pure Transformer or subquadratic models~\citep{lieber2024jambahybridtransformermambalanguage, ren2024sambasimplehybridstate, dong2024hymbahybridheadarchitecturesmall}. These architectures are particularly well-suited for the increased compute demands of inference-time scaling. Our work evaluates the inference-time scaling properties of both pure and hybrid subquadratic models. 

\subsection{Knowledge Distillation}
Knowledge distillation has proven effective in transferring capabilities from large teacher models to smaller, more efficient student models~\citep{hinton2015distillingknowledgeneuralnetwork}. In the context of LLMs, distillation is commonly used to compress a larger pre-trained LLM into a smaller version while maintaining core knowledge and functionality~\citep{gu2024minillmknowledgedistillationlarge, xu2024surveyknowledgedistillationlarge}.
Although larger models exhibit better reasoning and overall abilities due to the properties of scale~\citep{xu2025largereasoningmodelssurvey, wei2022emergentabilitieslargelanguage}, distillation has enabled smaller models to achieve strong reasoning performance~\citep{deepseekai2025deepseekr1incentivizingreasoningcapability, bespoke_stratos}. While most distillation efforts focus on within-architecture transfer (e.g., Transformer to Transformer), recent work has explored cross-architecture distillation. Pretrained Transformers have been successfully distilled into recurrent architectures such as RNNs~\citep{kasai2021finetuningpretrainedtransformersrnns, mercat2024linearizinglargelanguagemodels}, linear attention~\citep{zhang2024hedgehogporcupineexpressive}, convolutions~\citep{ralambomihanta2024scavenginghyenadistillingtransformers}, and SSMs~\citep{bick2024transformersssmsdistillingquadratic,wang2025mamballamadistillingaccelerating}. Whether strong reasoning can be distilled across architectures remains an open question.
\section{Distilling Student Reasoners}
\label{sec:method:distill}

\begin{figure*}[h]
  \centering
  \begin{subfigure}[b]{0.4\linewidth}
  \includegraphics[width=\textwidth,clip]{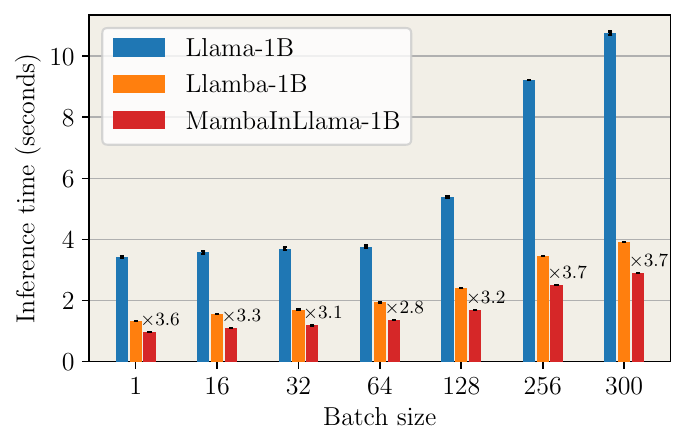}
  \caption{1B Models.}
  \label{fig:speedup-1b}
  \end{subfigure}\hspace{4em}
  \begin{subfigure}[b]{0.39\linewidth}
  \includegraphics[width=\textwidth,clip]{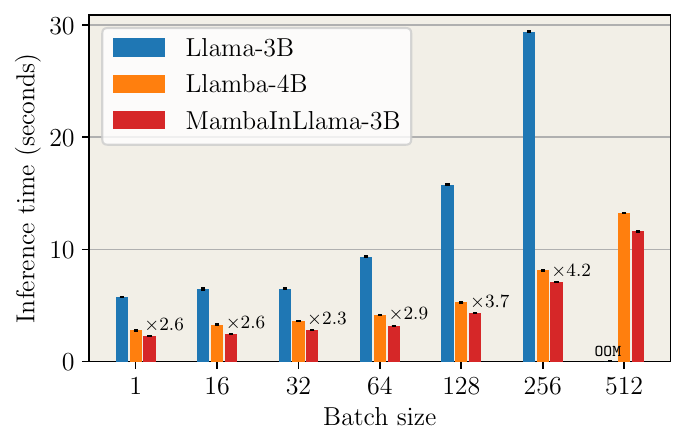}
  \caption{3B Models.}
  \label{fig:speedup-3b}
  \end{subfigure}
\caption{\textbf{Faster generation of distilled models.}
In \textbf{(a)} and \textbf{(b)}, we show the inference time measured for the baseline Llama models as well as our distilled Llamba (pure Mamba) and MambaInLlama (hybrid) models at the 1B and 3B scale. We denote the speedup for our MambaInLlama model at each batch size. We use prompts of $512$ tokens and measure the time required to generate $512$ tokens. The times measured do not include the prefilling of the prompt. Overall, distilled models can generate tokens much faster with the speedup being greater for larger batch sizes. Moreover, our distilled models are more memory efficient, as shown in \textbf{(b)}, using a batch size of $512$ yields an Out of Memory (OOM) error for Llama 3B, but not for our models. To obtain $512$ completions with Llama-3B, the two batches of $256$, result in an inference time of $58.8$s. In comparison, our MambaInLlama model would take $11.6$s, a speedup of $\times5.1$.   
 }
\label{fig:speedup}
\end{figure*}

\begin{figure}[h]
  \centering
  \begin{subfigure}[b]{0.48\linewidth}
  \includegraphics[width=\textwidth,clip]{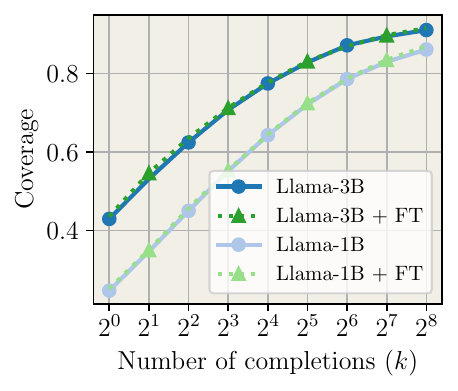}
  \caption{Coverage.}
  \end{subfigure}
  \begin{subfigure}[b]{0.48\linewidth}
  \includegraphics[width=\textwidth,clip]{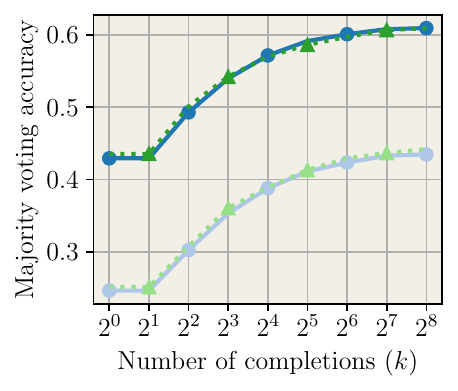}
  \caption{Majority voting.}
  \end{subfigure}
\caption{\textbf{Negligible effect of finetuning Llama baselines on distillation dataset.} As our distillation dataset includes math content, we also finetune Llama models on the distillation dataset OpenMathInstruct-2. Those models are marked with the "+FT" prefix. We plot the coverage \textbf{(a)} and majority voting accuracies \textbf{(b)} as a function of the number of completions. We observe that finetuning Llama models on the distillation dataset has a negligible effect on those metrics. 
 }
\label{fig:llama-ft}
\end{figure}

\begin{algorithm}[tb]
    \small
    \label{alg:group_ssm}
    \begin{algorithmic}[1]
        \STATE \textbf{Shapes:} $B$ - Batch, $L$ - Length, $D$ - embed size, \\  
        \hspace{1cm} $N = D  / \text{Heads}$, $N'$ - expand  
        \STATE \textbf{Input:} $\vo_t$: (B,  D)
        \STATE \textbf{Output:} output: (B, D) 
        \STATE \textbf{New Params:} MLP, $\rmA$  
        \FOR{ each head $\rmW^k, \rmW^q, \rmW^v, \rmW^o : (N, D)$ \\
             \hspace{1cm} expanding grouped KVs}
            \STATE \textbf{Head Parameter:} $\rmA : (N, N')$
            \STATE \text{for all positions $t$:}
            \STATE $\ \vx_t : (B, N) \gets \rmW^V \vo_t$
            \STATE $\ \rmB_t : (B, N) \gets \rmW^K \vo_t$ 
            \STATE $\ \rmC_t : (B, N) \gets \rmW^Q \vo_t$
            \STATE $\ \Delta_t : (B, N') \gets \text{MLP}(\vx_t)$
            \STATE $\overline{\rmA}_{1:T}, \overline{\rmB}_{1:T}, \overline{\rmC}_{1:T}: (B, N, N') \gets \textsc{Disc}(\rmA, \rmB, \rmC, \Delta)$
            \STATE $\vy \gets \textsc{LinearRNN}(\overline{\rmA}, \overline{\rmB}, \overline{\rmC}, \vx)$
            \STATE output $\gets \text{output} + \rmW^{O\top} \vy$
        \ENDFOR{}
        \STATE \textbf{return} output
    \end{algorithmic}
    \caption{Initializing MambaInLlama from Llama}
\end{algorithm}

In this section, we describe how we distilled Llama models into pure Mamba and hybrid architectures. We refer to our pure Mamba models as \textbf{Llamba}, and our hybrid models as \textbf{MambaInLlama}. We distill both our hybrid and pure Mamba models using Llama 3.2-1B-Instruct and Llama 3.2-3B-Instruct from the Llama family of models~\citep{grattafiori2024llama3herdmodels}. 

\subsection{Distilling into Llamba}

\textbf{Distillation method.} In order to distill pure Mamba models, we modify the MOHAWK distillation procedure introduced by~\citet{bick2024transformersssmsdistillingquadratic}.
MOHAWK is composed of three stages: 1) matrix orientation, 2) hidden state alignment, and 3) weight transfer and knowledge distillation. Stage 1 (matrix orientation) aligns the Mamba-2 model's SSM matrix mixer~\citep{dao2024transformersssmsgeneralizedmodels} with the teacher's self-attention matrix by minimizing the distance between the two matrices. Stage 2 (hidden state alignment) matches the student and teacher's layers' hidden state outputs. Both of these stages are run independently across layers to prevent previous optimization gaps from propagation through the model. This is done by setting the input of the student layer to be that of the previous teacher layer's output.
Stage 3 (weight transfer and knowledge distillation) transfers the remaining, unoptimized parameters, e.g., MLPs, embeddings, and norms, and finetunes the complete end-to-end student model using a distillation loss on the student and teacher logits~\citep{hinton2015distillingknowledgeneuralnetwork}. We deviate from the original MOHAWK paper by transferring the MLP weights and norms of each teacher decoder layer to the student and training those parameters as well during Stage 2. This stems from the architectural differences between Phi~\citep{li2023textbooksneediiphi15} (MLP and self-attention in parallel) and Llama~\citep{grattafiori2024llama3herdmodels} (sequential self-attention and MLP). Stage 3 remains the same with fewer weights transferred.

\textbf{Experimental details.} Our pure Mamba-distilled models, Llamba-1B and Llamba-4B, dubbed after \citet{llamba}, which uses a similar methodology and the same teacher models, are distilled from their respective teacher models using our adjusted MOHAWK distillation approach with only 8 billion tokens total each. Following \citet{bick2024transformersssmsdistillingquadratic}, we use a variant of Mamba-2 that converts the SSM head structure to multi-head (compared to the original multi-value and Llama's grouped-query structure) and converts the sequence mixer to entirely discrete-time. The post-convolution activation and pre-output projection normalization are also removed.
The 8B token distillation dataset is composed of 4B tokens from FineMath-4+~\citep{lozhkov2024finemath}, allocated as 1B and 3B to Stages 1 and 2 respectively, and 4B tokens from OpenMathInstruct-2~\citep{toshniwal2024openmathinstruct2acceleratingaimath} used in Stage 3, which is the only stage in which we apply the chat template to the inputs. Unlike for our hybrid models, we find that computing the loss on both the assistant output and user prompt improves model performance over just the assistant output. All three distillation stages use the AdamW optimizer with $\beta=(0.9, 0.95)$ and weight decay of $0.1$ and a Warmup-Stable-Decay (WSD) scheduler with $10$\% warmup and $10$\% decay~\citep{hu2024minicpmunveilingpotentialsmall} at a 2048 context length. In Stages 1 and 2, we set the learning rate to $1\times 10^{-4}$, while in Stage 3, it is set to $1\times 10^{-5}$. The hyperparameters are the same for the 1B and 4B distillation runs. Our final Llamba-1B model has 16 layers of our Mamba-2 variant with a state size of $64$ and multi-head pattern of 32 heads and state size of $64$. Likewise, our Llamba-4B utilizes the same pattern with $24$ heads and state size of $128$ for $28$ layers. We note that our pure Mamba-2 models are slightly larger than their Transformer counterparts due to the pattern conversion from grouped-query to multi-head and additional parameters found within the Mamba-2 layer, e.g., gating.

\subsection{Distilling into MambaInLlama}

\textbf{Distillation method.} We follow two separate directions for distillation.
For the hybrid models, we modify the protocol proposed by ~\citet{wang2025mamballamadistillingaccelerating} in order to distill some specific capabilities. These techniques have been shown to be effective for hybrid architectures.
The Mamba-in-Llama framework ~\citep{wang2025mamballamadistillingaccelerating} introduces a method for distilling hybrid Transformer-Mamba models by reusing weights from the attention layers. In the distillation process shown in Alg~1, the linear projections for $\rmQ$, $\rmK$, $\rmV$ and $\rmO$ are initialized using the corresponding linear projections for $\rmC$, $\rmB$, $\rmX$ and $\rmO$ respectively. The only additional learned parameters in the new layers are the sampling rate $\Delta$ and the dynamic $\rmA$. These new parameters will control the constructed Mamba through
the discretization function. Specifically, we take $\Delta \in \R^{N'}$ to discretize $\rmB_t, \rmC_t \in \R^{N \times 1}$ and obtain $\overline{\rmB}_t, \overline{\rmC}_t \in \R^{N'\times N\times 1}$ as shown in Alg~1.
We directly reuse the MLP layers. Differently from \citet{wang2025mamballamadistillingaccelerating}, we replace the attention layers with Mamba layers in a single round and finetune the whole model. For distillation, we employ token-level KL divergence. The full probability distribution of the student model, $p(\cdot; \theta)$, is trained to align with the full distribution of the teacher model, $p(\cdot; \theta_T)$, by minimizing the KL divergence across all possible next tokens at position $t$. Different from \citep{wang2025mamballamadistillingaccelerating}, we use the reverse KL divergence, $ D_{\text{KL}}(p(\cdot; \theta) \parallel p(\cdot; \theta_T))$ instead of the forward KL divergence as the loss function,
since reverse KL behaves more like mode-seeking and better mimics the peak values. And we find that it yields better results empirically.
We adopt the Mamba-1 architecture for our hybrid models, as~\citet{lieber2024jambahybridtransformermambalanguage,junxiongdaniele2024mambainllama,dong2024hymbahybridheadarchitecturesmall} demonstrates that using Mamba-1 in a hybrid architecture yields better results, especially for challenging reasoning tasks.
 
\paragraph{Experimental details.} Our hybrid Mamba models, named MambaInLlama-1B (with $4$ attention layers in $16$ total layers) and MambaInLlama-3B (with $6$ attention layers in $26$ total layers), are distilled with 8B tokens from OpenMathInstruct-2 \citep{toshniwal2024openmathinstruct2acceleratingaimath}. We apply the Llama chat template, mask the user prompt, and compute the loss only over the tokens generated in the assistant's output. Thus, the total number of supervised tokens is reduced to roughly 7B. To speed up training, we use data packing to merge different sequences into a single one until we reach the maximum sequence length which is set to $8192$. We use the AdamW optimizer with learning rate $2\times 10^{-5}$, $\beta=(0.9, 0.95)$ and a weight decay of $0.1$. The hyperparameters are the same for the 1B and 3B models. 
In Mamba layers, we set the SSM state size to $16$. Consequently, the number of SSM groups after expansion is $2048/16 = 128$ for the 1B model and $3072/16 = 192$ for the 3B model.

\textbf{Remarks on the distillation dataset.}
We finetune the Llama teacher models on the same data used during distillation to avoid our models from potentially gaining an unfair advantage. The results, reported in Figure~\ref{fig:llama-ft}, show that the continuous training of the base model on the distillation data mix has a negligible effect on performance. Moreover, we find that the selection of data used during distillation has a significant impact on the final capabilities of the distilled models. Switching the Stage 3 dataset in Llamba-1B from OpenMathInstruct-2 to OpenHermes-2.5~\citep{openhermes-2p5} decreased greedy decoding accuracy on MATH (acc@1) by more than 10 percentage points.

\textbf{Lack of correlation between reasoning and general benchmarks.} We also highlight the lack of correlation between common multiple choice-based benchmarks and mathematical reasoning, as the OpenHermes variant of Llamba-1B outperforms the final Llamba-1B by more than 5 points on MMLU~\citep{hendryckstest2021} and $0.5$ point on ARC-Challenge~\citep{clark2018thinksolvedquestionanswering}. Moreover, analyzing the acc@1 performance of Llamba-1B and Llamba-4B on MATH after each of the three stages, we see that the sharp increase in reasoning ability between Stage 2 and Stage 3 is not reflected in general knowledge benchmarks (Fig.~\ref{fig:mohawk-reasoning}).

\subsection{Improving performance after distillation}
We show that it is possible to improve the accuracy and coverage of our distilled models by performing some supervised fine-tuning (SFT) after distillation. 
This is inspired by \citet{wang2025mamballamadistillingaccelerating}, where SFT is an integral part of the distillation protocol.
Starting from the distilled MambainLlama-1B and 3B, we fine-tune the models for two epochs using 8 billion tokens from OpenMathInstruct-2.
The distilled models achieve impressive performance both in coverage and accuracy, even surpassing the original Llama models. This is illustrated in Figure~\ref{fig:sftdpo}.

\section{Scaling Inference Time Compute}

\begin{figure*}[t!]
  \centering
  \begin{subfigure}[b]{0.575\linewidth}
  \includegraphics[width=\textwidth,clip]{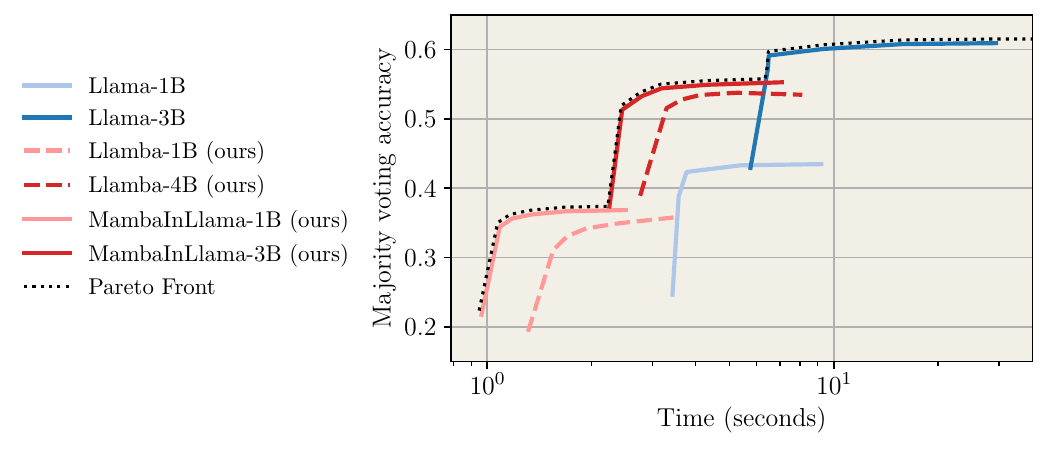}
  \caption{Majority voting.}
  \label{fig:acc-math-a}
  \end{subfigure}
  \begin{subfigure}[b]{0.38\linewidth}
  \includegraphics[width=\textwidth,clip]{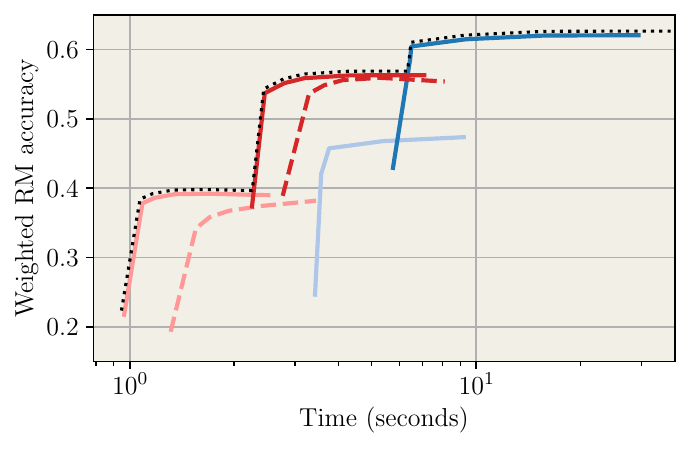}
  \caption{Weighted Best-of-N.}
  \label{fig:acc-math-b}
  \end{subfigure}
\caption{\textbf{Distilled models provide better accuracies on MATH for most time budgets.} Figures are similar to Figure~\ref{fig:coverage-math-a}.
In \textbf{(a)} and \textbf{(b)}, we show the majority-voting accuracy and the weight Best-of-N accuracy (the selected answer is the one with the highest sum of reward model scores, as introduced in \citet{beeching2024scalingtesttimecompute}), for different time budgets. Similarly to Figure~\ref{fig:coverage-math-a}, we observe how the higher throughput of our distilled models allows them to push the Pareto front for most time budgets. Interestingly, when comparing models of a given size, Llama models are better for larger time budgets. However, looking at both model sizes together reveals that larger distilled models can compensate for the lower accuracies of smaller models. As a result, the Pareto front is defined by our hybrid models. While Llama-3B is still more efficient for a large time budget, one could imagine distilling a larger subquadratic model that generates quicker nonetheless.
}
\label{fig:acc-math}
\end{figure*}

\label{sec:method:scale}

\begin{figure*}[h]
  \centering
  \begin{subfigure}[b]{0.575\linewidth}
  \includegraphics[width=\textwidth,clip]{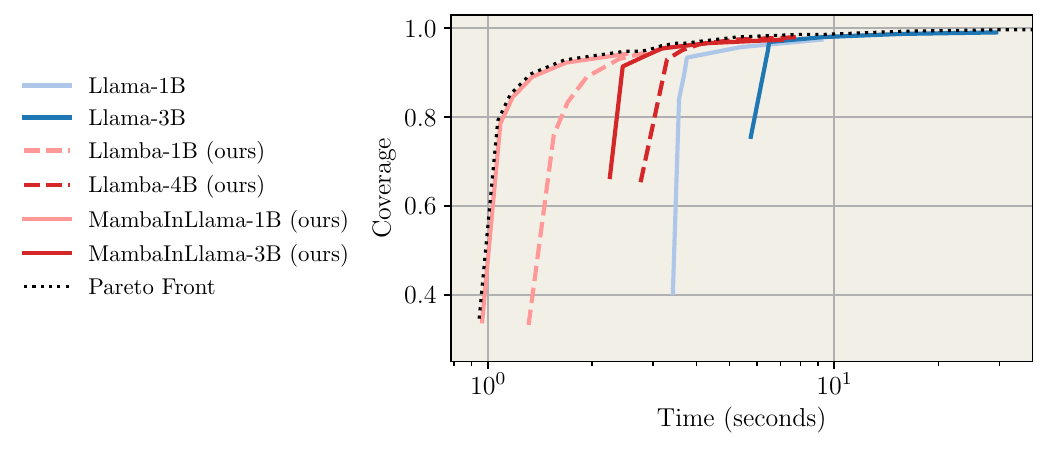}
  \caption{Scaling with time.}
  \label{fig:coverage-gsm8k-b}
  \end{subfigure}
  \begin{subfigure}[b]{0.38\linewidth}
  \includegraphics[width=\textwidth,clip]{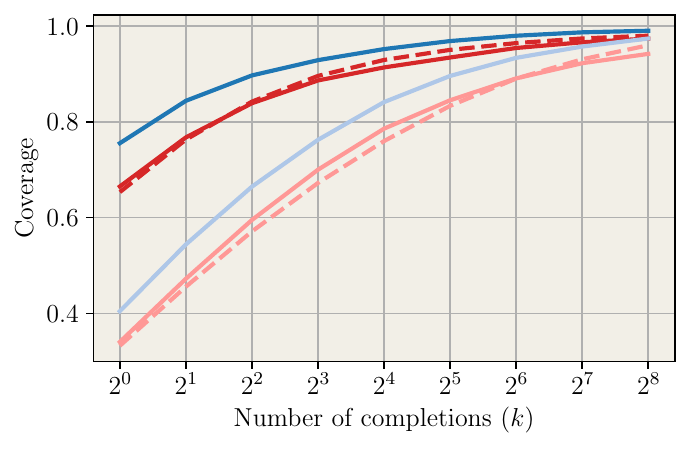}
  \caption{Scaling with number of completions.}
  \label{fig:coverage-gsm8k-a}
  \end{subfigure}
\caption{\textbf{Distilled models have better coverage on GSM8K for most time budgets.}
Observations are similar to Figure~\ref{fig:coverage-math}, despite a small degradation in coverage per number of completions, the better throughput of distilled models pushes the Pareto front forward. Hybrid models are more efficient for most time budgets.  
}
\label{fig:coverage-gsm8k}
\end{figure*}

\begin{figure*}[h]
  \centering
  \begin{subfigure}[b]{0.575\linewidth}
  \includegraphics[width=\textwidth,clip]{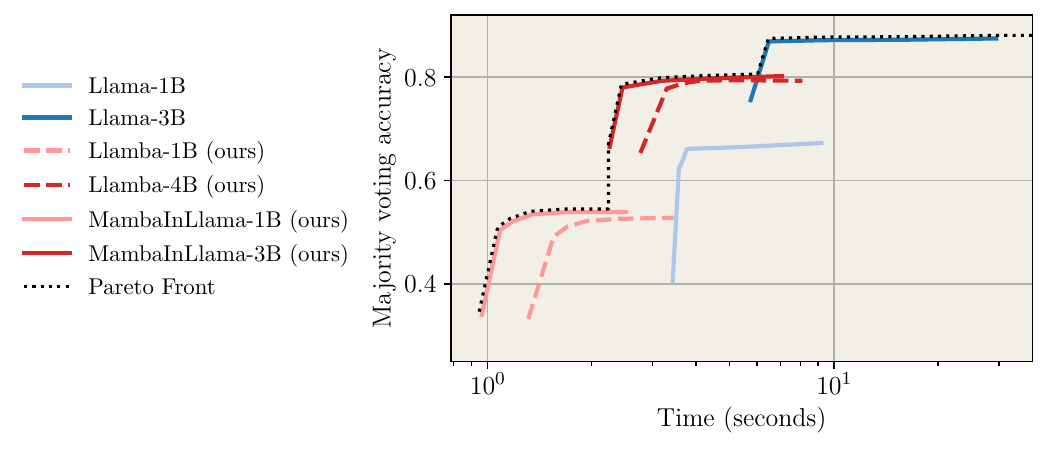}
  \caption{Majority voting.}
  \label{fig:acc-gsm8k-a}
  \end{subfigure}
  \begin{subfigure}[b]{0.38\linewidth}
  \includegraphics[width=\textwidth,clip]{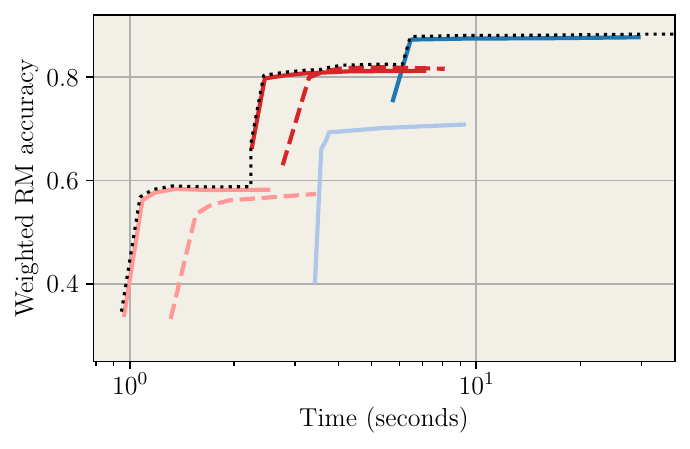}
  \caption{Weighted Best-of-N.}
  \label{fig:acc-gsm8k-b}
  \end{subfigure}
\caption{\textbf{Distilled models provide better accuracies on GSM8K for most time budgets.}
Observations are similar to Figure~\ref{fig:acc-math}. The higher throughput of our distilled models allows them to push the Pareto front for most time budgets.}
\label{fig:acc-gsm8k}
\end{figure*}

We scale test-time compute using our distilled models by generating multiple CoTs to solve a set of math problems.
The system prompt (Figure~\ref{fig:dataset_prompt}) contains instructions on how to properly format the response.
The model outputs are parsed to extract the final solution, which is then compared to the ground truth. This approach enables us to evaluate the model's performance in generating correct solutions across multiple attempts.
Moreover, the results demonstrate that the models are able to retain their instruction following ability after distillation.

\paragraph{Evaluation metrics.} We evaluate our model using two primary metrics: coverage and accuracy.
In domains like coding and formal proofs, where answers can be automatically verified, coverage directly translates to improved performance and has been widely adopted~\citep{chen2021evaluatinglargelanguagemodels, brown2024largelanguagemonkeysscaling}.
Coverage is commonly referred to as the pass@k metric, where $k$ denotes the number of samples per problem~\citep{chen2021evaluatinglargelanguagemodels, brown2024largelanguagemonkeysscaling}.
This metric estimates the probability that at least one correct solution exists among the $k$ samples.
To reduce the variance when calculating coverage, we
adopt the unbiased estimation formula from~\citet{chen2021evaluatinglargelanguagemodels}. Specifically, we generate $N\geq k$ total samples per task. The probability that a correct solution exists among a pool of $k$ generated samples can then be determined given the total number of correct solutions $C_i$ for each task.

\[
\text{pass@k} = \frac{1}{\# \text{ of problems}} \sum_{i=1}^{\# \text{ of problems}} \left( 1 - \frac{\binom{N - C_i}{k}}{\binom{N}{k}} \right)
\]

We implement this formula using a numerically stable approach as suggested by~\citet{chen2021evaluatinglargelanguagemodels}(see Appendix~\ref{app:pass_k}).

For accuracy, we use multiple aggregation strategies.
Majority voting, or self-consistency decoding~\citep{wang2023selfconsistencyimproveschainthought}
is the most straightforward method to aggregate responses and compute an accuracy score.
A more refined strategy involves using a trained verifier to select the best response (we call this approach Best-of-N).

As our verifier, we utilize a reward model trained using process supervision, where the model receives feedback on each step of the reasoning process.
Inspired by~\citet{snell2024scalingllmtesttimecompute}, we utilize a Llama-3.1 8B-based reward model to score the solutions for Best-of-N. As PRMs produce a cumulative sequence of step-level scores per solution, we perform a reduction over the steps to obtain a single solution-level score which we use for answer selection. 
Following~\citet{snell2024scalingllmtesttimecompute}, we use the final score in all the steps as the score for Best-of-N.

While Best-of-N simply selects the generation with the highest reward, weighted Best-of-N aggregates the rewards of samples that share the same response and returns the solution with the highest combined score. In this work, we chose to report weighted Best-of-N, since we often found it to be superior to Best-of-N, most likely due to it identifying highly-rated but also common solutions.
We focus on the performance of the models for a fixed compute budget.
Inspired by the scaling laws for inference-time compute ~\citep{brown2024largelanguagemonkeysscaling}, we speculate that having a model that can scale to very large number of completions can be very effective, even more so than increasing model size by a large amount ~\citep{snell2024scalingllmtesttimecompute}.

\section{Results}
\label{sec:results}

We (i) measure the inference speedup of our distilled models (\S~\ref{subsec:speedup}), and (ii) show that this speedup can result in better scaling for a given inference time budget (\S~\ref{subsec:scaling}).

\subsection{Inference time results}
\label{subsec:speedup}

\textbf{Experimental protocol.} In order to measure the inference time and throughput of our distilled models, we focus on creating a realistic setup that matches the prompt and CoTs lengths that we find for MATH and GSM8K. We compare the runtime of a Llama 3.2 architecture against our distilled models. For Llama, we use FlashAttention2 \citep{dao2023flashattention2fasterattentionbetter} and torch compile. The implementations of our models rely on the standard Mamba implementation. Given a varying batch size, we consider the tasks of generating $512$ tokens from a $512$ token prompt, which reflect the length found in the evaluation datasets. The prefilling time to process the prompt is not included in the benchmark, as it may depend on the redundancy of a given prompt within the batch. In fact, in our setting, we are only interested in the time to generate the multiple completions given one prompt. Our benchmark is done on a single NVIDIA H100 GPU, and averaged results over multiple runs are shown in Figure~\ref{fig:speedup}.

\textbf{Distilled models are significantly faster.} Results in Figure~\ref{fig:speedup} show our distilled models are up to $\times3.7$ and $\times4.2$ faster than their respective Llama 1B and 3B baselines. Moreover, MambaInLlama and Llamba are more memory efficient and thus can run larger batches. In Figure~\ref{fig:speedup-3b}, our models can accommodate batches of $512$ while Llama-3B returns an out-of-memory error. We also notice that MambaInLlama models are slightly faster than Llamba. We speculate that this is because MambaInLlama has a smaller SSM state size of $16$, while Llamba uses a larger SSM state size of $64$. Additionally, Llamba has larger projections because of its multi-head structure.

\textbf{Remark.} In our speed experiments, both the prompt and the generation lengths are relatively short compared to many other domains. 
When evaluating multi-turn conversations, for example, where distilled Mamba architectures are already shown to excel~\citep{wang2025mamballamadistillingaccelerating}, it is clear that the length of the context and of the CoTs can be significantly larger than what has been considered in our experiments. Such longer sequences would significantly increase the throughput advantage of our models. Unfortunately, it is not yet clear how to evaluate and exploit test-time compute for more subjective, conversational tasks.

\textbf{Limitations.} We try to ensure fair speed comparisons between the model types by utilizing roughly equivalent implementations: standard implementation of Mamba-1/2 and FlashAttention-based self-attention with Pytorch compilation. However, it is worth noting that there exist optimizations for current Transformer architectures, such as memory management improvements~\citep{kwon2023efficient}, that enable faster generation; similar boosts are currently not readily available for alternative architectures. Deeper optimizations for each architecture are beyond the scope of this work.

\begin{figure*}[t!]
    \centering
    \begin{tcolorbox}[]
        \textbf{MATH System Prompt:} \small\textit{Solve the following math problem efficiently and clearly:\textbackslash n\textbackslash n- For simple problems (2 steps or fewer):\textbackslash nProvide a concise solution with minimal explanation.\textbackslash n\textbackslash n- For complex problems (3 steps or more):\textbackslash nUse this step-by-step format:\textbackslash n\textbackslash n\#\# Step 1: [Concise description]\textbackslash n[Brief explanation and calculations]\textbackslash n\textbackslash n\#\# Step 2: [Concise description]\textbackslash n[Brief explanation and calculations]\textbackslash n\textbackslash n...\textbackslash n\textbackslash nRegardless of the approach, always conclude with:\textbackslash n\textbackslash nTherefore, the final answer is: \$\textbackslash boxed\{answer\}\$. I hope it is correct.\textbackslash n\textbackslash nWhere [answer] is just the final number or expression that solves the problem.
    }

    \end{tcolorbox}
    \begin{tcolorbox}[]
            \textbf{GSM8K System Prompt:} \small\textit{\textbackslash n\textbackslash nGiven the following problem, reason and give a final answer to the problem.\textbackslash nYour response should end with "The final answer is [answer]" where [answer] is the response to the problem.\textbackslash nProblem:
    }
    \end{tcolorbox}
    
    \caption{\textbf{System prompts for MATH and GSM8K.} 
    System prompts passed to all models (pure and hybrid distilled models and Llama baseline) to be used within the chat template when generating solutions to questions in the MATH and GSM8K datasets.}
    \label{fig:dataset_prompt}
\end{figure*}

\subsection{Results on reasoning tasks} 
\label{subsec:scaling}

\textbf{Experimental protocol.} We benchmark the teacher models (Llama-3.2 1B-Instruct and 3B-Instruct) and the distilled Mamba students (MambaInLlama-1B, MambaInLlama-3B, Llamba-1B, Llamba-4B) on the MATH-500 subset of the MATH dataset~\citep{lightman2023letsverifystepstep, hendrycks2021measuringmathematicalproblemsolving} and a randomly selected 500-sample subset of GSM8K~\citep{cobbe2021trainingverifierssolvemath}. We evaluate performance based on coverage and accuracy with majority voting and weighted Best-of-N (\S~\ref{sec:method:scale}). We sample responses from the distilled models with temperatures $T=0.6,0.8$ and $\mathsf{top\_k=-1}$ to consider all tokens. For each response, we sample up to $2048$ tokens. For the distilled models, we find that $T=0.6$ and $T=0.8$ are ideal for the 1B and 3B scale, respectively, and subsequent results are obtained using these temperatures. The official Llama evaluation system prompt is used for the MATH dataset, while the original prompt is kept for GSM8K as displayed in Figure~\ref{fig:dataset_prompt}. For our process reward model, we utilize a base Llama3.1-8B-Instruct model trained on Mistral-generated data~\citep{xiong2024rlhflowmath}.

\begin{figure*}[h]
  \centering
  \begin{subfigure}[b]{0.363\linewidth}
  \includegraphics[width=\textwidth,clip]{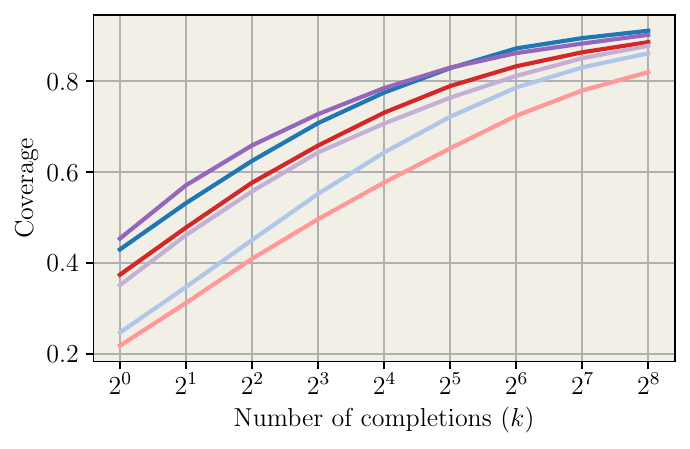}
  \caption{Coverage.}
  \label{fig:sft-coverage-1b}
  \end{subfigure}\hspace{2em}
  \begin{subfigure}[b]{0.59\linewidth}
  \includegraphics[width=\textwidth,clip]{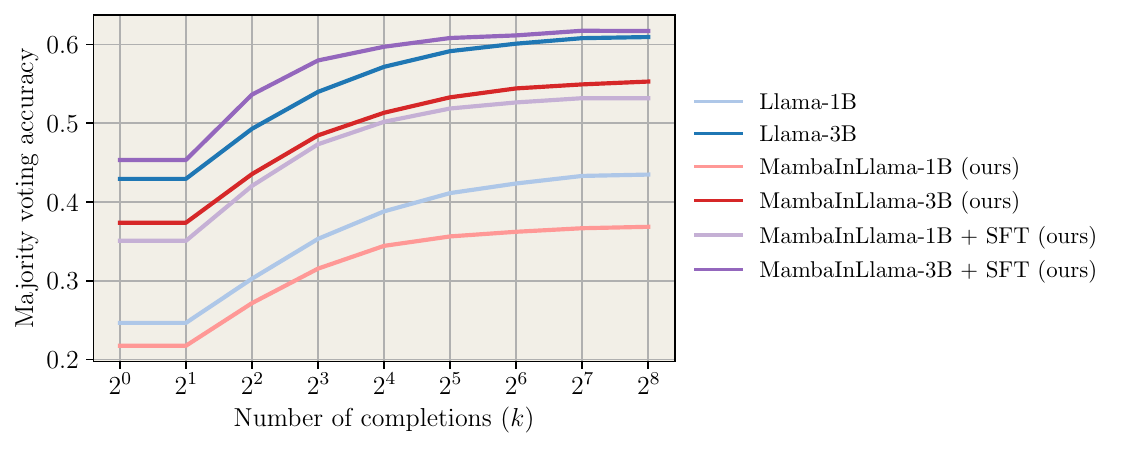}
  \caption{Majority voting.}
  \label{fig:sft-acc-1b}
  \end{subfigure}
\caption{\textbf{SFT can further improve distilled models.}
In \textbf{(a)}, and \textbf{(b)}, we show the coverage and majority voting accuracies for MambaInLlama distilled models. We show how coverage and majority voting accuracies can be further improved through Supervised Fine-Tuning (SFT). The improvement is especially striking for our 1B model. 
 }
\label{fig:sftdpo}
\end{figure*}

\textbf{Distilled Models can Cover like Teachers.} When observing the scaling of coverage as the number of generation $k$ increases in Figure~\ref{fig:coverage-math-b} (resp. Fig.~\ref{fig:coverage-gsm8k-b} for GSM8K), we see distilled models closely matching the coverage of their teachers. Only a small degradation is observed. When we plot the coverage as a function of the time budget, by associating each coverage measurement to the time required to generate the associated number of completions (see Fig.~\ref{fig:coverage-math-a}), we find that our distilled models are exceptional in their ability to generate correct answers fast. By generating many more completions for the same time budget, the overall Pareto front for coverage in both MATH and GSM8K (Fig.~\ref{fig:coverage-math-a},~\ref{fig:coverage-gsm8k-a}) is heavily dominated by our distilled models where both pure and hybrid Mamba reasoners are able to achieve the same degree of coverage in nearly half the time of their respective teachers. Given a sufficiently strong reward model or verifiable solutions, those coverage scores would in large parts translate to accuracy.

\textbf{Distilled Models Achieve Competitive Accuracy Under Fixed Time.}
Training with distillation and utilizing subquadratic architectures---which are less expressive than Transformers---can degrade model quality. However, we find that this is a worthy trade-off when comparing performance under fixed time budgets in Figure~\ref{fig:acc-gsm8k} and Figure~\ref{fig:acc-math}.

Similarly to coverage, the lighter and faster batch inference of the distilled models, allowing for more generations, results in a better accuracy/time Pareto front at several completion scales. Interestingly, while comparing models of similar sizes indicates that larger time budgets are dominated by the teacher models, we observe that the larger distilled model can provide better accuracy than the smaller baseline while still being faster. For example, while Llama-1B provides better accuracy than MambaInLlama-1B for larger time budgets, MambaInLlama-3B takes over where MambaInLlama-1B left off, providing a better accuracy than Llama-1B for inference time. Similarly, we conjecture that it would be possible to distill a larger baseline model that would outperform Llama-3B, providing better accuracy for a given time budget.

\textbf{Larger Students Faster and Better than Smaller Teachers.} The core driving force behind the growing interest in subquadratic models is their computational efficiency. Such properties enable our distilled models of larger sizes (3B scale) to generate samples faster than even a smaller Transformer (1B). Our MambaInLlama-3B and Llamba-4B models outperform the slower Llama-1B baseline on coverage and accuracy while being faster. These inference speedups, rooted in the underlying architecture, allow larger and more capable models to be used in time-constrained environments, despite their increased number of parameters.
\textbf{Smaller models have great coverage.} When focusing on coverage, we observe in Figure~\ref{fig:coverage-math-a} and Figure~\ref{fig:coverage-gsm8k-a} that most of the Pareto front is occupied by 1B models. Contrasting those results with the majority voting accuracy results in Figure~\ref{fig:acc-math-a} and Figure~\ref{fig:acc-gsm8k-a} where the gap between 1B and 3B models is much more significant. An interpretation is that, while smaller models have the ability to generate the correct answer, the probability of generating it within a constrained number of samples increases with the model size. This finding has implications in choosing model size for tasks involving formal language where the answer is easily verifiable, such as coding and mathematical proofs. In those applications, coverage matters most, and smaller models may be preferred given their better time/coverage efficiency compared to larger models.

\textbf{SFT improves the models significantly}
Following distillation, which establishes a strong foundation for our models, we observe that additional supervised fine-tuning (SFT) significantly enhances their performance, as illustrated in Figure~\ref{fig:sftdpo}. This suggests that while distillation effectively transfers knowledge from the teacher model, SFT further refines and aligns the model's capabilities, enabling it to become highly competitive. Our results indicate that by leveraging both distillation and SFT, subquadratic architectures can match or even surpass their Transformer teachers in absolute performance on reasoning tasks.
\section{Conclusion}
\label{sec:conclusion}
In our work, we investigate whether lower-complexity models can leverage their superior generation throughput to outperform similarly sized Transformers under a fixed computational budget. We focus on reasoning tasks where we can scale test-time compute to improve performance.
Through extensive experimentation, we distill both pure and hybrid Mamba models at the 1B and 3B scales and evaluate their reasoning capabilities on mathematical reasoning benchmarks, where the ability of subquadratic models to quickly generate many completions enables them to take advantage of their scaling properties when increasing inference compute. 
When fixing memory and/or compute, our models achieve better coverage and accuracy for most time budgets compared to their Transformer, teacher counterparts.
These findings highlight the potential of Mamba and other attention alternatives as strong substitutes to Transformers for tasks that benefit from scalable inference compute.

We hope this work inspires future work in pretraining subquadratic reasoners and further exploring their inference scaling properties. More research is required to determine the best way to distill reasoning capabilities across architectures, as performance remains highly sensitive to both data and distillation techniques. Moreover, since our distilled models demonstrate exceptional coverage, developing better reward models to better identify correct answers can close the accuracy gap.
Finally, further investigation into scaling inference compute for conversational and subjective tasks would open to a scenario where lighter subquadratic models can achieve larger gains in performance and speed

\section{Acknowledgements}
We thank Cartesia and TogetherAI for their support in providing compute for this project. KL is supported by funding from the Bosch Center for Artificial Intelligence.

\bibliography{paper}

\begin{thebibliography}{55}
\providecommand{\natexlab}[1]{#1}
\providecommand{\url}[1]{\texttt{#1}}
\expandafter\ifx\csname urlstyle\endcsname\relax
  \providecommand{\doi}[1]{doi: #1}\else
  \providecommand{\doi}{doi: \begingroup \urlstyle{rm}\Url}\fi

\bibitem[Beck et~al.(2024)Beck, Pöppel, Spanring, Auer, Prudnikova, Kopp,
  Klambauer, Brandstetter, and Hochreiter]{beck2024xlstmextendedlongshortterm}
Beck, M., Pöppel, K., Spanring, M., Auer, A., Prudnikova, O., Kopp, M.,
  Klambauer, G., Brandstetter, J., and Hochreiter, S.
\newblock xlstm: Extended long short-term memory, 2024.
\newblock URL \url{https://arxiv.org/abs/2405.04517}.

\bibitem[Beeching et~al.(2024)Beeching, Tunstall, and
  Rush]{beeching2024scalingtesttimecompute}
Beeching, E., Tunstall, L., and Rush, S.
\newblock Scaling test-time compute with open models, 2024.
\newblock URL
  \url{https://huggingface.co/spaces/HuggingFaceH4/blogpost-scaling-test-time-compute}.

\bibitem[Bick et~al.(2024)Bick, Li, Xing, Kolter, and
  Gu]{bick2024transformersssmsdistillingquadratic}
Bick, A., Li, K.~Y., Xing, E.~P., Kolter, J.~Z., and Gu, A.
\newblock Transformers to ssms: Distilling quadratic knowledge to subquadratic
  models, 2024.
\newblock URL \url{https://arxiv.org/abs/2408.10189}.

\bibitem[Bick et~al.(2025)Bick, Katsch, Sohoni, Desai, and Gu]{llamba}
Bick, A., Katsch, T., Sohoni, N., Desai, A., and Gu, A.
\newblock Llamba: Scaling distilled recurrent models for efficient language
  processing, 2025.
\newblock URL \url{https://arxiv.org/abs/2502.14458}.

\bibitem[Brown et~al.(2024)Brown, Juravsky, Ehrlich, Clark, Le, Ré, and
  Mirhoseini]{brown2024largelanguagemonkeysscaling}
Brown, B., Juravsky, J., Ehrlich, R., Clark, R., Le, Q.~V., Ré, C., and
  Mirhoseini, A.
\newblock Large language monkeys: Scaling inference compute with repeated
  sampling, 2024.
\newblock URL \url{https://arxiv.org/abs/2407.21787}.

\bibitem[Chen et~al.(2021)Chen, Tworek, Jun, Yuan, de~Oliveira~Pinto, Kaplan,
  Edwards, Burda, Joseph, Brockman, Ray, Puri, Krueger, Petrov, Khlaaf, Sastry,
  Mishkin, Chan, Gray, and et. al.]{chen2021evaluatinglargelanguagemodels}
Chen, M., Tworek, J., Jun, H., Yuan, Q., de~Oliveira~Pinto, H.~P., Kaplan, J.,
  Edwards, H., Burda, Y., Joseph, N., Brockman, G., Ray, A., Puri, R., Krueger,
  G., Petrov, M., Khlaaf, H., Sastry, G., Mishkin, P., Chan, B., Gray, S., and
  et. al.
\newblock Evaluating large language models trained on code, 2021.
\newblock URL \url{https://arxiv.org/abs/2107.03374}.

\bibitem[Clark et~al.(2018)Clark, Cowhey, Etzioni, Khot, Sabharwal, Schoenick,
  and Tafjord]{clark2018thinksolvedquestionanswering}
Clark, P., Cowhey, I., Etzioni, O., Khot, T., Sabharwal, A., Schoenick, C., and
  Tafjord, O.
\newblock Think you have solved question answering? try arc, the ai2 reasoning
  challenge, 2018.
\newblock URL \url{https://arxiv.org/abs/1803.05457}.

\bibitem[Cobbe et~al.(2021)Cobbe, Kosaraju, Bavarian, Chen, Jun, Kaiser,
  Plappert, Tworek, Hilton, Nakano, Hesse, and
  Schulman]{cobbe2021trainingverifierssolvemath}
Cobbe, K., Kosaraju, V., Bavarian, M., Chen, M., Jun, H., Kaiser, L., Plappert,
  M., Tworek, J., Hilton, J., Nakano, R., Hesse, C., and Schulman, J.
\newblock Training verifiers to solve math word problems, 2021.
\newblock URL \url{https://arxiv.org/abs/2110.14168}.

\bibitem[Dao(2023)]{dao2023flashattention2fasterattentionbetter}
Dao, T.
\newblock Flashattention-2: Faster attention with better parallelism and work
  partitioning, 2023.
\newblock URL \url{https://arxiv.org/abs/2307.08691}.

\bibitem[Dao \& Gu(2024)Dao and Gu]{dao2024transformersssmsgeneralizedmodels}
Dao, T. and Gu, A.
\newblock Transformers are ssms: Generalized models and efficient algorithms
  through structured state space duality, 2024.
\newblock URL \url{https://arxiv.org/abs/2405.21060}.

\bibitem[DeepSeek-AI et~al.(2025)DeepSeek-AI, Guo, Yang, Zhang, Song, Zhang,
  Xu, Zhu, Ma, Wang, Bi, Zhang, Yu, Wu, Wu, Gou, Shao, Li, Gao, and et.
  al.]{deepseekai2025deepseekr1incentivizingreasoningcapability}
DeepSeek-AI, Guo, D., Yang, D., Zhang, H., Song, J., Zhang, R., Xu, R., Zhu,
  Q., Ma, S., Wang, P., Bi, X., Zhang, X., Yu, X., Wu, Y., Wu, Z.~F., Gou, Z.,
  Shao, Z., Li, Z., Gao, Z., and et. al.
\newblock Deepseek-r1: Incentivizing reasoning capability in llms via
  reinforcement learning, 2025.
\newblock URL \url{https://arxiv.org/abs/2501.12948}.

\bibitem[Dong et~al.(2024)Dong, Fu, Diao, Byeon, Chen, Mahabaleshwarkar, Liu,
  Keirsbilck, Chen, Suhara, Lin, Kautz, and
  Molchanov]{dong2024hymbahybridheadarchitecturesmall}
Dong, X., Fu, Y., Diao, S., Byeon, W., Chen, Z., Mahabaleshwarkar, A.~S., Liu,
  S.-Y., Keirsbilck, M.~V., Chen, M.-H., Suhara, Y., Lin, Y., Kautz, J., and
  Molchanov, P.
\newblock Hymba: A hybrid-head architecture for small language models, 2024.
\newblock URL \url{https://arxiv.org/abs/2411.13676}.

\bibitem[Goyal et~al.(2024)Goyal, Ji, Rawat, Menon, Kumar, and
  Nagarajan]{goyal2024thinkspeaktraininglanguage}
Goyal, S., Ji, Z., Rawat, A.~S., Menon, A.~K., Kumar, S., and Nagarajan, V.
\newblock Think before you speak: Training language models with pause tokens,
  2024.
\newblock URL \url{https://arxiv.org/abs/2310.02226}.

\bibitem[Grattafiori et~al.(2024)Grattafiori, Dubey, Jauhri, Pandey, Kadian,
  Al-Dahle, Letman, Mathur, Schelten, Vaughan, Yang, Fan, Goyal, Hartshorn,
  Yang, Mitra, Sravankumar, Korenev, Hinsvark, and et.
  al.]{grattafiori2024llama3herdmodels}
Grattafiori, A., Dubey, A., Jauhri, A., Pandey, A., Kadian, A., Al-Dahle, A.,
  Letman, A., Mathur, A., Schelten, A., Vaughan, A., Yang, A., Fan, A., Goyal,
  A., Hartshorn, A., Yang, A., Mitra, A., Sravankumar, A., Korenev, A.,
  Hinsvark, A., and et. al.
\newblock The llama 3 herd of models, 2024.
\newblock URL \url{https://arxiv.org/abs/2407.21783}.

\bibitem[Gu \& Dao(2024)Gu and Dao]{gu2024mambalineartimesequencemodeling}
Gu, A. and Dao, T.
\newblock Mamba: Linear-time sequence modeling with selective state spaces,
  2024.
\newblock URL \url{https://arxiv.org/abs/2312.00752}.

\bibitem[Gu et~al.(2022)Gu, Goel, and
  Ré]{gu2022efficientlymodelinglongsequences}
Gu, A., Goel, K., and Ré, C.
\newblock Efficiently modeling long sequences with structured state spaces,
  2022.
\newblock URL \url{https://arxiv.org/abs/2111.00396}.

\bibitem[Gu et~al.(2024)Gu, Dong, Wei, and
  Huang]{gu2024minillmknowledgedistillationlarge}
Gu, Y., Dong, L., Wei, F., and Huang, M.
\newblock Minillm: Knowledge distillation of large language models, 2024.
\newblock URL \url{https://arxiv.org/abs/2306.08543}.

\bibitem[Hendrycks et~al.(2021{\natexlab{a}})Hendrycks, Burns, Basart, Zou,
  Mazeika, Song, and Steinhardt]{hendryckstest2021}
Hendrycks, D., Burns, C., Basart, S., Zou, A., Mazeika, M., Song, D., and
  Steinhardt, J.
\newblock Measuring massive multitask language understanding.
\newblock \emph{Proceedings of the International Conference on Learning
  Representations (ICLR)}, 2021{\natexlab{a}}.

\bibitem[Hendrycks et~al.(2021{\natexlab{b}})Hendrycks, Burns, Kadavath, Arora,
  Basart, Tang, Song, and
  Steinhardt]{hendrycks2021measuringmathematicalproblemsolving}
Hendrycks, D., Burns, C., Kadavath, S., Arora, A., Basart, S., Tang, E., Song,
  D., and Steinhardt, J.
\newblock Measuring mathematical problem solving with the math dataset,
  2021{\natexlab{b}}.
\newblock URL \url{https://arxiv.org/abs/2103.03874}.

\bibitem[Hinton et~al.(2015)Hinton, Vinyals, and
  Dean]{hinton2015distillingknowledgeneuralnetwork}
Hinton, G., Vinyals, O., and Dean, J.
\newblock Distilling the knowledge in a neural network, 2015.
\newblock URL \url{https://arxiv.org/abs/1503.02531}.

\bibitem[Hu et~al.(2024)Hu, Tu, Han, He, Cui, Long, Zheng, Fang, Huang, Zhao,
  Zhang, Thai, Zhang, Wang, Yao, Zhao, Zhou, Cai, Zhai, and et.
  al.]{hu2024minicpmunveilingpotentialsmall}
Hu, S., Tu, Y., Han, X., He, C., Cui, G., Long, X., Zheng, Z., Fang, Y., Huang,
  Y., Zhao, W., Zhang, X., Thai, Z.~L., Zhang, K., Wang, C., Yao, Y., Zhao, C.,
  Zhou, J., Cai, J., Zhai, Z., and et. al.
\newblock Minicpm: Unveiling the potential of small language models with
  scalable training strategies, 2024.
\newblock URL \url{https://arxiv.org/abs/2404.06395}.

\bibitem[Kasai et~al.(2021)Kasai, Peng, Zhang, Yogatama, Ilharco, Pappas, Mao,
  Chen, and Smith]{kasai2021finetuningpretrainedtransformersrnns}
Kasai, J., Peng, H., Zhang, Y., Yogatama, D., Ilharco, G., Pappas, N., Mao, Y.,
  Chen, W., and Smith, N.~A.
\newblock Finetuning pretrained transformers into rnns, 2021.
\newblock URL \url{https://arxiv.org/abs/2103.13076}.

\bibitem[Katharopoulos et~al.(2020)Katharopoulos, Vyas, Pappas, and
  Fleuret]{katharopoulos2020transformersrnnsfastautoregressive}
Katharopoulos, A., Vyas, A., Pappas, N., and Fleuret, F.
\newblock Transformers are rnns: Fast autoregressive transformers with linear
  attention, 2020.
\newblock URL \url{https://arxiv.org/abs/2006.16236}.

\bibitem[Kwon et~al.(2023)Kwon, Li, Zhuang, Sheng, Zheng, Yu, Gonzalez, Zhang,
  and Stoica]{kwon2023efficient}
Kwon, W., Li, Z., Zhuang, S., Sheng, Y., Zheng, L., Yu, C.~H., Gonzalez, J.~E.,
  Zhang, H., and Stoica, I.
\newblock Efficient memory management for large language model serving with
  pagedattention.
\newblock In \emph{Proceedings of the ACM SIGOPS 29th Symposium on Operating
  Systems Principles}, 2023.

\bibitem[Labs(2025)]{bespoke_stratos}
Labs, B.
\newblock Bespoke-stratos: The unreasonable effectiveness of reasoning
  distillation.
\newblock
  www.bespokelabs.ai/blog/bespoke-stratos-the-unreasonable-effectiveness-of-reasoning-distillation,
  2025.
\newblock Accessed: 2025-01-22.

\bibitem[Li et~al.(2023)Li, Bubeck, Eldan, Giorno, Gunasekar, and
  Lee]{li2023textbooksneediiphi15}
Li, Y., Bubeck, S., Eldan, R., Giorno, A.~D., Gunasekar, S., and Lee, Y.~T.
\newblock Textbooks are all you need ii: phi-1.5 technical report, 2023.
\newblock URL \url{https://arxiv.org/abs/2309.05463}.

\bibitem[Lieber et~al.(2024)Lieber, Lenz, Bata, Cohen, Osin, Dalmedigos,
  Safahi, Meirom, Belinkov, Shalev-Shwartz, Abend, Alon, Asida, Bergman,
  Glozman, Gokhman, Manevich, Ratner, Rozen, Shwartz, Zusman, and
  Shoham]{lieber2024jambahybridtransformermambalanguage}
Lieber, O., Lenz, B., Bata, H., Cohen, G., Osin, J., Dalmedigos, I., Safahi,
  E., Meirom, S., Belinkov, Y., Shalev-Shwartz, S., Abend, O., Alon, R., Asida,
  T., Bergman, A., Glozman, R., Gokhman, M., Manevich, A., Ratner, N., Rozen,
  N., Shwartz, E., Zusman, M., and Shoham, Y.
\newblock Jamba: A hybrid transformer-mamba language model, 2024.
\newblock URL \url{https://arxiv.org/abs/2403.19887}.

\bibitem[Lightman et~al.(2023)Lightman, Kosaraju, Burda, Edwards, Baker, Lee,
  Leike, Schulman, Sutskever, and Cobbe]{lightman2023letsverifystepstep}
Lightman, H., Kosaraju, V., Burda, Y., Edwards, H., Baker, B., Lee, T., Leike,
  J., Schulman, J., Sutskever, I., and Cobbe, K.
\newblock Let's verify step by step, 2023.
\newblock URL \url{https://arxiv.org/abs/2305.20050}.

\bibitem[Lozhkov et~al.(2024)Lozhkov, Ben~Allal, Bakouch, von Werra, and
  Wolf]{lozhkov2024finemath}
Lozhkov, A., Ben~Allal, L., Bakouch, E., von Werra, L., and Wolf, T.
\newblock Finemath: the finest collection of mathematical content, 2024.
\newblock URL \url{https://huggingface.co/datasets/HuggingFaceTB/finemath}.

\bibitem[Luo et~al.(2024)Luo, Liu, Liu, Phatale, Guo, Lara, Li, Shu, Zhu, Meng,
  Sun, and Rastogi]{luo2024improvemathematicalreasoninglanguage}
Luo, L., Liu, Y., Liu, R., Phatale, S., Guo, M., Lara, H., Li, Y., Shu, L.,
  Zhu, Y., Meng, L., Sun, J., and Rastogi, A.
\newblock Improve mathematical reasoning in language models by automated
  process supervision, 2024.
\newblock URL \url{https://arxiv.org/abs/2406.06592}.

\bibitem[Mercat et~al.(2024)Mercat, Vasiljevic, Keh, Arora, Dave, Gaidon, and
  Kollar]{mercat2024linearizinglargelanguagemodels}
Mercat, J., Vasiljevic, I., Keh, S., Arora, K., Dave, A., Gaidon, A., and
  Kollar, T.
\newblock Linearizing large language models, 2024.
\newblock URL \url{https://arxiv.org/abs/2405.06640}.

\bibitem[Peng et~al.(2023)Peng, Alcaide, Anthony, Albalak, Arcadinho, Biderman,
  Cao, Cheng, Chung, Grella, GV, He, Hou, Lin, Kazienko, Kocon, Kong, Koptyra,
  Lau, and et. al.]{peng2023rwkvreinventingrnnstransformer}
Peng, B., Alcaide, E., Anthony, Q., Albalak, A., Arcadinho, S., Biderman, S.,
  Cao, H., Cheng, X., Chung, M., Grella, M., GV, K.~K., He, X., Hou, H., Lin,
  J., Kazienko, P., Kocon, J., Kong, J., Koptyra, B., Lau, H., and et. al.
\newblock Rwkv: Reinventing rnns for the transformer era, 2023.
\newblock URL \url{https://arxiv.org/abs/2305.13048}.

\bibitem[Pfau et~al.(2024)Pfau, Merrill, and Bowman]{pfau2024letsthinkdotdot}
Pfau, J., Merrill, W., and Bowman, S.~R.
\newblock Let's think dot by dot: Hidden computation in transformer language
  models, 2024.
\newblock URL \url{https://arxiv.org/abs/2404.15758}.

\bibitem[Qwen et~al.(2025)Qwen, :, Yang, Yang, Zhang, Hui, Zheng, Yu, Li, Liu,
  Huang, Wei, Lin, Yang, Tu, Zhang, Yang, Yang, Zhou, Lin, Dang, and et.
  al.]{qwen2025qwen25technicalreport}
Qwen, :, Yang, A., Yang, B., Zhang, B., Hui, B., Zheng, B., Yu, B., Li, C.,
  Liu, D., Huang, F., Wei, H., Lin, H., Yang, J., Tu, J., Zhang, J., Yang, J.,
  Yang, J., Zhou, J., Lin, J., Dang, K., and et. al.
\newblock Qwen2.5 technical report, 2025.
\newblock URL \url{https://arxiv.org/abs/2412.15115}.

\bibitem[Ralambomihanta et~al.(2024)Ralambomihanta, Mohammadzadeh, Islam,
  Jabbour, and Liang]{ralambomihanta2024scavenginghyenadistillingtransformers}
Ralambomihanta, T.~R., Mohammadzadeh, S., Islam, M. S.~N., Jabbour, W., and
  Liang, L.
\newblock Scavenging hyena: Distilling transformers into long convolution
  models, 2024.
\newblock URL \url{https://arxiv.org/abs/2401.17574}.

\bibitem[Ren et~al.(2024)Ren, Liu, Lu, Shen, Liang, and
  Chen]{ren2024sambasimplehybridstate}
Ren, L., Liu, Y., Lu, Y., Shen, Y., Liang, C., and Chen, W.
\newblock Samba: Simple hybrid state space models for efficient unlimited
  context language modeling, 2024.
\newblock URL \url{https://arxiv.org/abs/2406.07522}.

\bibitem[Renze \& Guven(2024)Renze and
  Guven]{renze2024effectsamplingtemperatureproblem}
Renze, M. and Guven, E.
\newblock The effect of sampling temperature on problem solving in large
  language models, 2024.
\newblock URL \url{https://arxiv.org/abs/2402.05201}.

\bibitem[Snell et~al.(2024)Snell, Lee, Xu, and
  Kumar]{snell2024scalingllmtesttimecompute}
Snell, C., Lee, J., Xu, K., and Kumar, A.
\newblock Scaling llm test-time compute optimally can be more effective than
  scaling model parameters, 2024.
\newblock URL \url{https://arxiv.org/abs/2408.03314}.

\bibitem[Teknium(2023)]{openhermes-2p5}
Teknium.
\newblock Openhermes 2.5: An open dataset of synthetic data for generalist llm
  assistants, 2023.
\newblock URL \url{https://huggingface.co/datasets/teknium/OpenHermes-2.5}.

\bibitem[Toshniwal et~al.(2024)Toshniwal, Du, Moshkov, Kisacanin, Ayrapetyan,
  and Gitman]{toshniwal2024openmathinstruct2acceleratingaimath}
Toshniwal, S., Du, W., Moshkov, I., Kisacanin, B., Ayrapetyan, A., and Gitman,
  I.
\newblock Openmathinstruct-2: Accelerating ai for math with massive open-source
  instruction data, 2024.
\newblock URL \url{https://arxiv.org/abs/2410.01560}.

\bibitem[Uesato et~al.(2022)Uesato, Kushman, Kumar, Song, Siegel, Wang,
  Creswell, Irving, and Higgins]{uesato2022solvingmathwordproblems}
Uesato, J., Kushman, N., Kumar, R., Song, F., Siegel, N., Wang, L., Creswell,
  A., Irving, G., and Higgins, I.
\newblock Solving math word problems with process- and outcome-based feedback,
  2022.
\newblock URL \url{https://arxiv.org/abs/2211.14275}.

\bibitem[Wang et~al.(2024)Wang, Paliotta, May, Rush, and
  Dao]{junxiongdaniele2024mambainllama}
Wang, J., Paliotta, D., May, A., Rush, A.~M., and Dao, T.
\newblock The mamba in the llama: Distilling and accelerating hybrid models.
\newblock \emph{arXiv preprint arXiv:2408.15237}, 2024.

\bibitem[Wang et~al.(2025)Wang, Paliotta, May, Rush, and
  Dao]{wang2025mamballamadistillingaccelerating}
Wang, J., Paliotta, D., May, A., Rush, A., and Dao, T.
\newblock The mamba in the llama: Distilling and accelerating hybrid models.
\newblock \emph{Advances in Neural Information Processing Systems},
  37:\penalty0 62432--62457, 2025.

\bibitem[Wang et~al.(2023)Wang, Wei, Schuurmans, Le, Chi, Narang, Chowdhery,
  and Zhou]{wang2023selfconsistencyimproveschainthought}
Wang, X., Wei, J., Schuurmans, D., Le, Q., Chi, E., Narang, S., Chowdhery, A.,
  and Zhou, D.
\newblock Self-consistency improves chain of thought reasoning in language
  models, 2023.
\newblock URL \url{https://arxiv.org/abs/2203.11171}.

\bibitem[Wei et~al.(2022)Wei, Tay, Bommasani, Raffel, Zoph, Borgeaud, Yogatama,
  Bosma, Zhou, Metzler, Chi, Hashimoto, Vinyals, Liang, Dean, and
  Fedus]{wei2022emergentabilitieslargelanguage}
Wei, J., Tay, Y., Bommasani, R., Raffel, C., Zoph, B., Borgeaud, S., Yogatama,
  D., Bosma, M., Zhou, D., Metzler, D., Chi, E.~H., Hashimoto, T., Vinyals, O.,
  Liang, P., Dean, J., and Fedus, W.
\newblock Emergent abilities of large language models, 2022.
\newblock URL \url{https://arxiv.org/abs/2206.07682}.

\bibitem[Wei et~al.(2023)Wei, Wang, Schuurmans, Bosma, Ichter, Xia, Chi, Le,
  and Zhou]{wei2023chainofthoughtpromptingelicitsreasoning}
Wei, J., Wang, X., Schuurmans, D., Bosma, M., Ichter, B., Xia, F., Chi, E., Le,
  Q., and Zhou, D.
\newblock Chain-of-thought prompting elicits reasoning in large language
  models, 2023.
\newblock URL \url{https://arxiv.org/abs/2201.11903}.

\bibitem[Wu et~al.(2024)Wu, Sun, Li, Welleck, and
  Yang]{wu2024inferencescalinglawsempirical}
Wu, Y., Sun, Z., Li, S., Welleck, S., and Yang, Y.
\newblock Inference scaling laws: An empirical analysis of compute-optimal
  inference for problem-solving with language models, 2024.
\newblock URL \url{https://arxiv.org/abs/2408.00724}.

\bibitem[Xiong et~al.(2024)Xiong, Zhang, Jiang, and
  Zhang]{xiong2024rlhflowmath}
Xiong, W., Zhang, H., Jiang, N., and Zhang, T.
\newblock An implementation of generative prm.
\newblock \url{https://github.com/RLHFlow/RLHF-Reward-Modeling}, 2024.

\bibitem[Xu et~al.(2025)Xu, Hao, Zong, Wang, Zhang, Wang, Lan, Gong, Ouyang,
  Meng, Shao, Yan, Yang, Song, Ren, Hu, Li, Feng, Gao, and
  Li]{xu2025largereasoningmodelssurvey}
Xu, F., Hao, Q., Zong, Z., Wang, J., Zhang, Y., Wang, J., Lan, X., Gong, J.,
  Ouyang, T., Meng, F., Shao, C., Yan, Y., Yang, Q., Song, Y., Ren, S., Hu, X.,
  Li, Y., Feng, J., Gao, C., and Li, Y.
\newblock Towards large reasoning models: A survey of reinforced reasoning with
  large language models, 2025.
\newblock URL \url{https://arxiv.org/abs/2501.09686}.

\bibitem[Xu et~al.(2024)Xu, Li, Tao, Shen, Cheng, Li, Xu, Tao, and
  Zhou]{xu2024surveyknowledgedistillationlarge}
Xu, X., Li, M., Tao, C., Shen, T., Cheng, R., Li, J., Xu, C., Tao, D., and
  Zhou, T.
\newblock A survey on knowledge distillation of large language models, 2024.
\newblock URL \url{https://arxiv.org/abs/2402.13116}.

\bibitem[Yang et~al.(2024)Yang, Wang, Shen, Panda, and
  Kim]{yang2024gatedlinearattentiontransformers}
Yang, S., Wang, B., Shen, Y., Panda, R., and Kim, Y.
\newblock Gated linear attention transformers with hardware-efficient training,
  2024.
\newblock URL \url{https://arxiv.org/abs/2312.06635}.

\bibitem[Yao et~al.(2023)Yao, Yu, Zhao, Shafran, Griffiths, Cao, and
  Narasimhan]{yao2023treethoughtsdeliberateproblem}
Yao, S., Yu, D., Zhao, J., Shafran, I., Griffiths, T.~L., Cao, Y., and
  Narasimhan, K.
\newblock Tree of thoughts: Deliberate problem solving with large language
  models, 2023.
\newblock URL \url{https://arxiv.org/abs/2305.10601}.

\bibitem[Zhang et~al.(2024{\natexlab{a}})Zhang, Zhoubian, Hu, Yue, Dong, and
  Tang]{zhang2024restmctsllmselftrainingprocess}
Zhang, D., Zhoubian, S., Hu, Z., Yue, Y., Dong, Y., and Tang, J.
\newblock Rest-mcts*: Llm self-training via process reward guided tree search,
  2024{\natexlab{a}}.
\newblock URL \url{https://arxiv.org/abs/2406.03816}.

\bibitem[Zhang et~al.(2024{\natexlab{b}})Zhang, Bhatia, Kumbong, and
  Ré]{zhang2024hedgehogporcupineexpressive}
Zhang, M., Bhatia, K., Kumbong, H., and Ré, C.
\newblock The hedgehog \& the porcupine: Expressive linear attentions with
  softmax mimicry, 2024{\natexlab{b}}.
\newblock URL \url{https://arxiv.org/abs/2402.04347}.

\bibitem[Zhang et~al.(2023)Zhang, Chen, Shen, Ding, Tenenbaum, and
  Gan]{zhang2023planninglargelanguagemodels}
Zhang, S., Chen, Z., Shen, Y., Ding, M., Tenenbaum, J.~B., and Gan, C.
\newblock Planning with large language models for code generation, 2023.
\newblock URL \url{https://arxiv.org/abs/2303.05510}.

\end{thebibliography}
\bibliographystyle{icml2025}

\newpage
\appendix
\onecolumn
\section{Additional results}
\begin{figure*}[h]
  \centering
\begin{subfigure}[b]{0.38\linewidth}
  \includegraphics[width=\textwidth,clip]{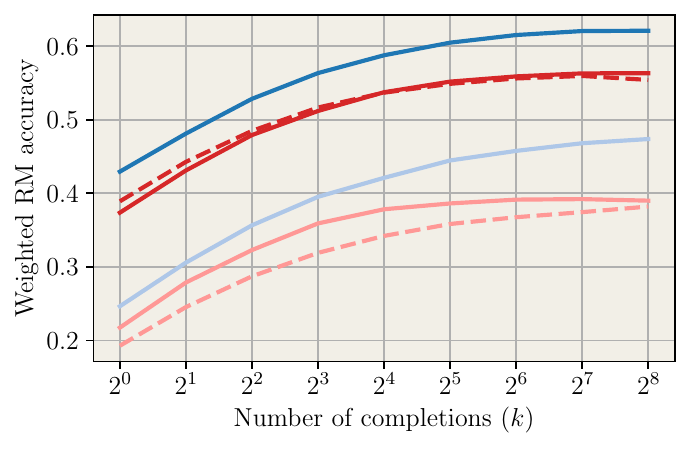}
  \caption{Weighted Best-of-N.}
  \end{subfigure}
  \begin{subfigure}[b]{0.575\linewidth}
  \includegraphics[width=\textwidth,clip]{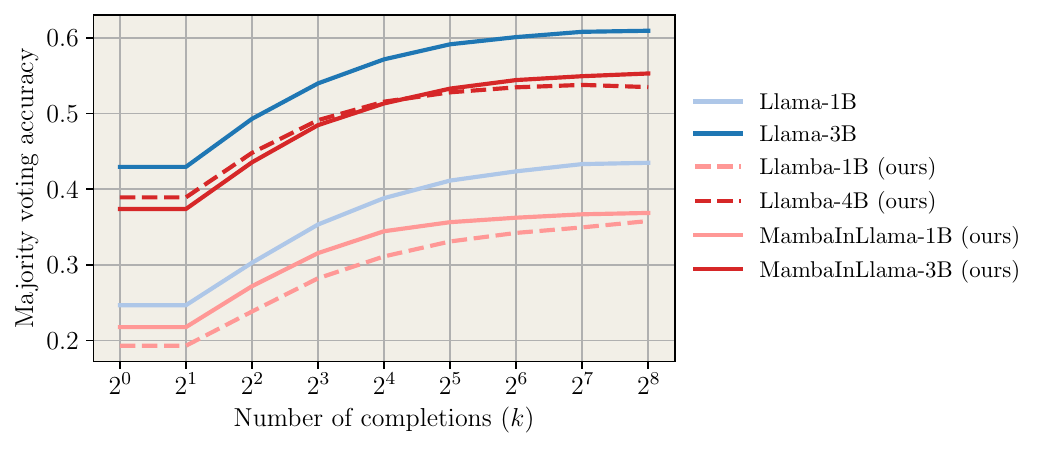}
  \caption{Majority voting.}
  \end{subfigure}
\caption{\textbf{Accuracy scores on MATH per completion without considering inference time.}
 }
\label{fig:w-acc-math}
\end{figure*}

\begin{figure*}[h]
  \centering
  \begin{subfigure}[b]{0.38\linewidth}
  \includegraphics[width=\textwidth,clip]{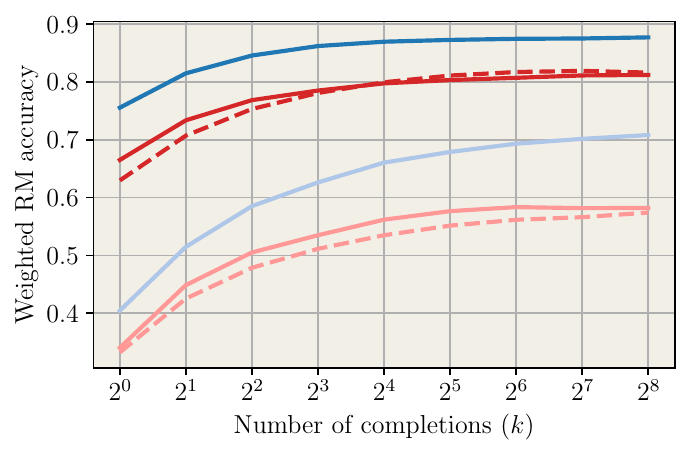}
  \caption{Weighted Best-of-N.}
  \end{subfigure}
  \begin{subfigure}[b]{0.575\linewidth}
  \includegraphics[width=\textwidth,clip]{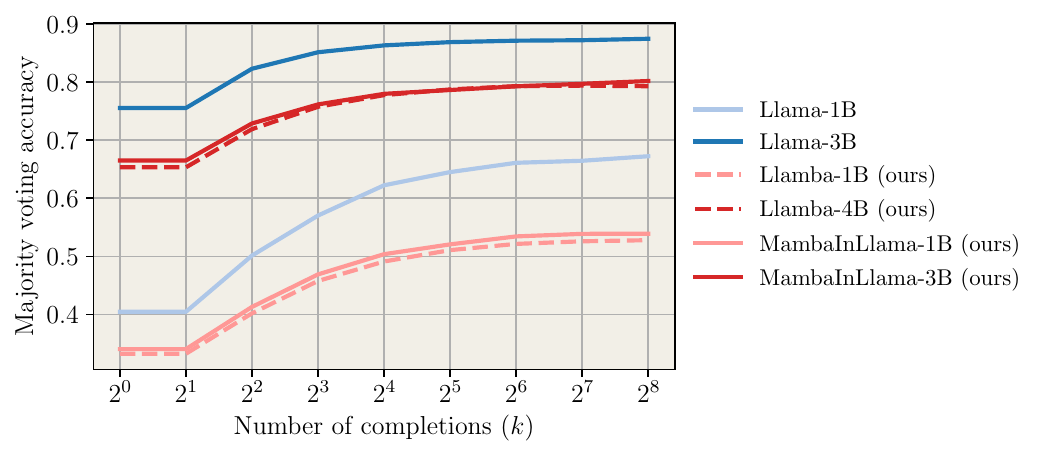}
  \caption{Majority voting.}
  \end{subfigure}
\caption{\textbf{Accuracy scores on GSM8K per completion without considering inference time.}
 }
\label{fig:w-acc-gsm8k}
\end{figure*}

Figures \ref{fig:w-acc-math} and \ref{fig:w-acc-gsm8k} show accuracy as a function of the number of completions.

\newpage
\section{Distillation loss curves}

\begin{figure}[h]
\centering
\includegraphics[width=0.8\textwidth,clip]{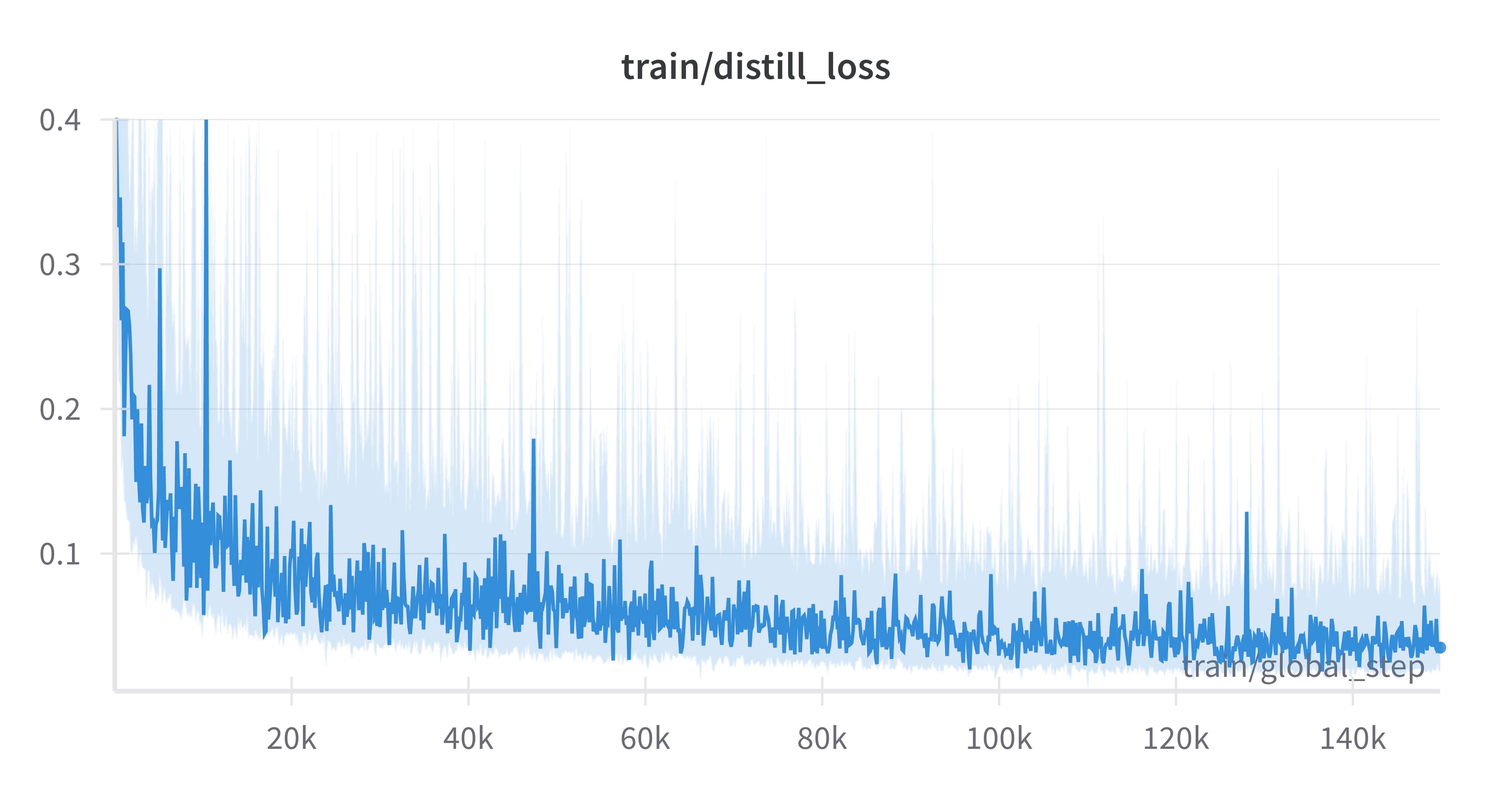}
\caption{\textbf{The distillation loss decreases steadily over time, showing a clear downward trend and converge at around 0.03.}}
\label{fig:mil3B_loss}
\end{figure}

Figure~\ref{fig:mil3B_loss} shows the distillation loss of MambaInLlama 3B when training on the \citet{toshniwal2024openmathinstruct2acceleratingaimath} dataset. We observed a stable decrease in loss until it converged to 0.03.

\section{Pass@k implementation}
\label{app:pass_k}
A numerically stable script for calculating an unbiased
estimate of pass@k, from~\citet{chen2021evaluatinglargelanguagemodels}.
\begin{lstlisting}[language=Python]
def pass_at_k(n, c, k):
    """
    :param n: total number of samples
    :param c: number of correct samples
    :param k: k in pass@$k$
    """
    if n - c < k: return 1.0
    return 1.0 - np.prod(1.0 - k /
                         np.arange(n - c + 1, n + 1))
\end{lstlisting}

\newpage
\section{MOHAWK (Pure Mamba) Distillation on Reasoning and General Performance}

\begin{figure}[h]
\centering
\includegraphics[width=\textwidth,clip]{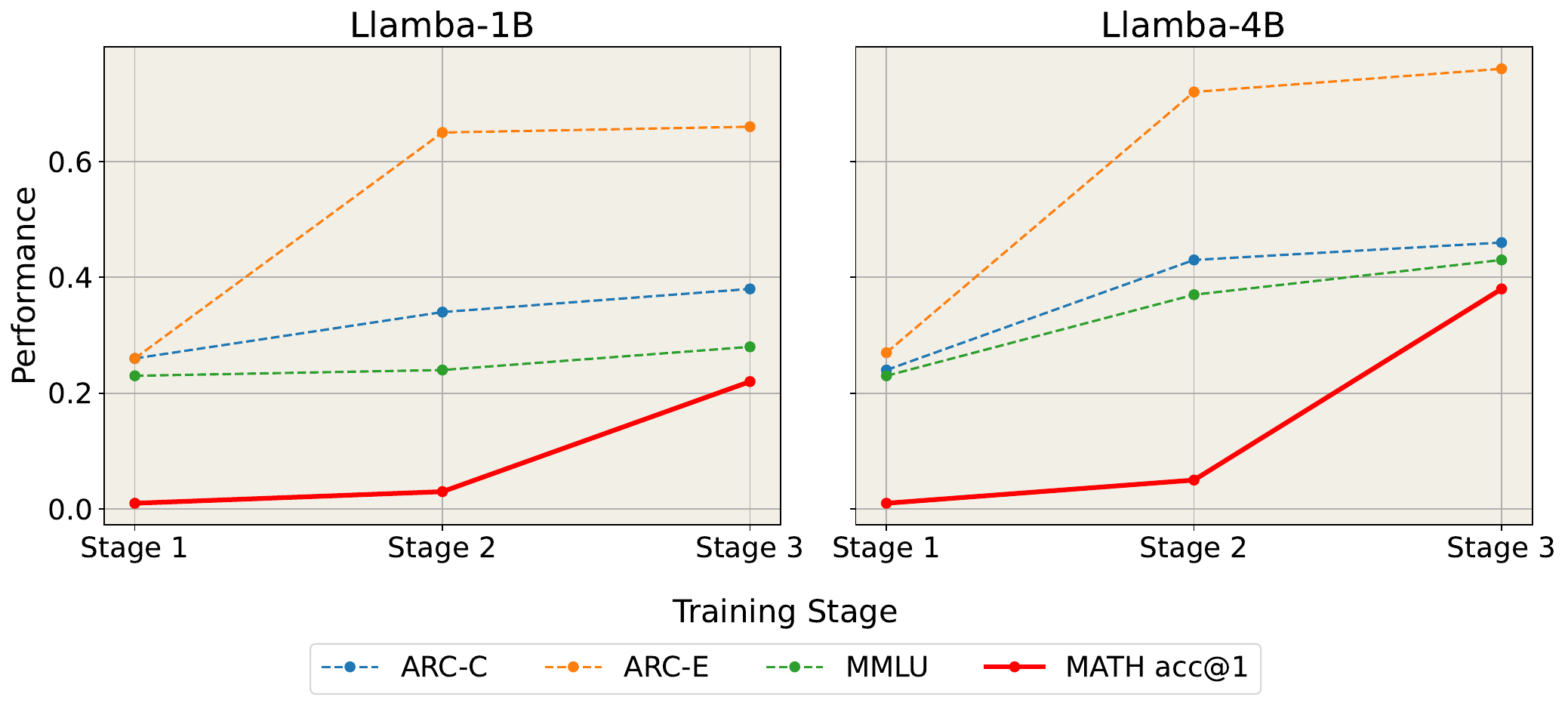}
\caption{\textbf{Reasoning and general performance of models after various stages of MOHAWK distillation.} Distilling reasoning capabilities, measured by accuracy using greedy decoding (acc@1) on the MATH benchmark, does not correlate with general multiple-choice reasoning benchmarks. The performance of Llamba-1B and 4B increase significantly only after the last stage of the MOHAWK distillation, but Stage 2 is able to significantly improve general benchmark performance.}
\label{fig:mohawk-reasoning}
\end{figure}

\end{document}